\documentclass[sigconf, nonacm, balance=false]{acmart}
\usepackage{colortbl}
\usepackage{multirow}
\usepackage{algorithm}
\usepackage{algpseudocode}
\usepackage{tabularx}
\usepackage{placeins}
\usepackage{float}
\usepackage[most]{tcolorbox}
\newcommand{\paperparagraph}[1]{\noindentparagraph{\normalfont\bfseries #1}}
\tcbuselibrary{listings}
\definecolor{sotashade}{RGB}{226,236,246}
\newcommand{\best}[1]{\cellcolor{sotashade}\textbf{#1}}
\newcommand{\ms}[2]{#1\,{\scriptsize\mdseries$\mathbin{\pm}$\,#2}}
\newlength{\nodecontentwidth}

\newtcblisting{promptbox}[1]{
  enhanced,
  listing only,
  title={#1},
  colback=black!1,
  colframe=black!38,
  colbacktitle=sotashade,
  coltitle=black,
  fonttitle=\bfseries\small,
  boxrule=0.45pt,
  arc=1mm,
  left=1.4mm,
  right=1.4mm,
  top=0.8mm,
  bottom=0.8mm,
  before skip=3pt,
  after skip=4pt,
  listing options={
    basicstyle=\footnotesize\ttfamily,
    breaklines=true,
    breakatwhitespace=false,
    columns=fullflexible,
    keepspaces=true,
    showstringspaces=false
  }
}

\AtBeginDocument{%
  }

\setcopyright{none}
\title{LLM-Enhanced Heterogeneous Graph Learning with Business Semantics for Financial Fraud Detection}

\author{Yixuan Chen}
\authornote{Both authors contributed equally to this research.}
\email{chenyx2123@mails.jlu.edu.cn}
\affiliation{%
  \institution{HKUST-GZ \& Jilin University}
  \city{Guangzhou}
  \country{China}
}

\author{Hongyu Zhan}
\authornotemark[1]
\email{hzhan701@connect.hkust-gz.edu.cn}
\affiliation{%
  \institution{The Hong Kong University of Science and Technology (Guangzhou)}
  \city{Guangzhou}
  \country{China}
}

\author{Jie Sheng}
\email{shengjie.sheng@ant-intl.com}
\affiliation{%
  \institution{Ant International}
  \city{Shanghai}
  \country{China}}

\author{Weiyu Han}
\email{hanweiyu.hwy@ant-intl.com}
\affiliation{%
  \institution{Ant International}
  \city{Shanghai}
  \country{China}}

\author{Shuai Chen}
\email{shuai.cs@ant-intl.com}
\affiliation{%
  \institution{Ant International}
  \city{Shanghai}
  \country{China}}

\author{Tianyi Zhang}
\email{zty113091@ant-intl.com}
\affiliation{%
  \institution{Ant International}
  \city{Shanghai}
  \country{China}}

\author{Xiao Tan}
\correspondingauthor
\email{alex.tx@ant-intl.com}
\affiliation{%
  \institution{Ant International}
  \city{Shanghai}
  \country{China}}

\author{Jun Xia}
\correspondingauthor
\email{junxia@hkust-gz.edu.cn}
\affiliation{%
  \institution{HKUST-GZ \& HKUST}
  \city{Hong Kong SAR}
  \country{China}
}

\begin{document}

% \begin{abstract}
% Financial fraud detection has evolved from classifying isolated transactions to reasoning over heterogeneous transaction graphs. Although heterogeneous graph neural networks preserve typed relations and structured attributes, they typically treat attributes as role-agnostic numerical inputs, leaving their risk-relevant business meaning implicit. Making this meaning explicit can provide complementary evidence for fraud detection on heterogeneous graphs. Large language models (LLMs) offer a way to derive such business semantics from node attributes, but their use in heterogeneous graph learning for financial fraud remains underexplored. To address this issue, we propose \textbf{SemHGT}, an LLM-enhanced framework that makes business semantics explicit through three stages: (1) Role-aware prompts elicit risk-related business descriptions from node attributes, and a text encoder maps them to semantic vectors for node enhancement; (2) Semantic similarity guides multi-round sampling of target-specific path instances under transaction-to-transaction schemas; (3) Schema-level tokens combining local path evidence and global neighbor summaries are processed with the target by a unified transformer. Across three benchmark datasets, SemHGT leads the evaluated baselines on most dataset--metric comparisons. Further analyses provide additional evidence for the effectiveness of SemHGT's design and clarify how its semantic modeling and evidence selection contribute to fraud detection. Code is available at \url{https://anonymous.4open.science/r/SemHGT-528A/}.
% \end{abstract}

\begin{abstract}
Financial fraud detection increasingly relies on heterogeneous transaction graphs, in which entities such as cards and merchants play distinct business roles and the risk meaning of an attribute depends on the entity's role: a high transaction volume may be routine for a merchant but unusual for a card. Heterogeneous graph neural networks preserve typed entities and relations but typically treat these attributes as role-agnostic inputs and leave their business meaning implicit. Meanwhile, most existing methods integrating large language models (LLMs) with graphs focus on homogeneous graphs or heterogeneous graphs with naturally available textual attributes, leaving their use on transaction graphs with structured attributes underexplored. To address this issue, we propose \textbf{SemHGT}, an LLM-enhanced framework that interprets structured attributes according to entity roles and makes the business semantics explicit in three stages: (1) role-aware prompts elicit risk-related business descriptions from node attributes, and a text encoder maps them to semantic vectors that enrich node representations; (2) semantic similarity guides multi-round sampling of target-specific path instances under transaction-to-transaction schemas; (3) schema-level tokens that combine local path evidence with global neighbor summaries are processed together with the target transaction by a unified transformer. Across three benchmark datasets, SemHGT leads the evaluated baselines on most dataset--metric comparisons. Further analyses provide additional evidence for the effectiveness of SemHGT's design and clarify how its semantic modeling and evidence selection contribute to fraud detection. Code is available at \url{https://anonymous.4open.science/r/SemHGT-528A/}.
\end{abstract}


\maketitle

\section{Introduction}
\label{sec:introduction}

Transactions arising from Web-based commerce are naturally represented as heterogeneous graphs~\cite{10.1145/3269206.3272010,10.1145/3366423.3380159,chen2026finfraudbench}. A transaction can be linked to different kinds of entities, such as cards, accounts, merchants, devices, and locations, through relations with distinct meanings~\cite{10.1145/3366423.3380159,chen2026finfraudbench,10.1145/3580305.3599803}. These typed entities and interactions provide complementary evidence for fraud detection: a target transaction may be suspicious because of the behavior of an associated entity~\cite{10.1145/3340531.3411903} or because it participates in a repeated relational pattern~\cite{10.1145/3442381.3449989}. By representing all entities with a single node type and all interactions with a single edge type, a homogeneous graph would obscure these roles and make it difficult to distinguish the evidence conveyed by different connections~\cite{wang2019heterogeneous,hu2020heterogeneous}.

Heterogeneous graph neural networks (HGNNs) offer a principled way to preserve this structure. Meta-path-based models aggregate information along predefined typed schemas~\cite{wang2019heterogeneous,fu2020magnn}, while meta-path-free models use type- and relation-aware message passing~\cite{10.1007/978-3-319-93417-4_38,hu2020heterogeneous,9737399}. More recent heterogeneous graph transformers further incorporate structural and relational encodings into attention~\cite{NIPS2017_3f5ee243,10.1145/3543507.3583493,10.24963/ijcai.2024/247}. Fraud-oriented graph neural networks (GNNs) address task-specific challenges by modeling temporal transaction behavior~\cite{xiang2023semi}, mitigating neighborhood camouflage through label-aware neighbor selection~\cite{10.1145/3340531.3411903}, addressing class imbalance with label-balanced sampling~\cite{10.1145/3442381.3449989}, and capturing attribute-association patterns across graph neighborhoods~\cite{duan2025global}. Despite these advances, real-world financial fraud detection presents a distinct semantic challenge. Transaction networks involve entities with different business roles and their node attributes consist primarily of structured numerical and categorical fields rather than natural-language text. Interpreting these attributes requires accounting for the role of each entity: high transaction volume, for example, may reflect routine merchant activity but unusual behavior for a card. Such role-dependent risk implications are not explicit in the raw attributes.

Large language models (LLMs) have been used in graph learning to provide textual features and auxiliary semantic supervision~\cite{ICLR2024_1766d75b,gao2025bootstrapping}. However, most existing LLM-enhanced graph methods focus on homogeneous graphs or heterogeneous graphs with naturally available textual attributes, leaving heterogeneous transaction graphs with predominantly structured attributes less explored. Recent fraud-detection studies employ graph-conditioned prompting~\cite{li2026dgp} and LLM-guided contrastive objectives~\cite{ou2026targeting}, but the use of role-aware business semantics in heterogeneous fraud detection remains limited. Such semantics can enrich node representations with risk-related information and guide target-specific sampling of local evidence under typed relations. Their joint use in these two stages remains underexplored. To address this gap, we propose the \textbf{Sem}antics-enhanced \textbf{H}eterogeneous \textbf{G}raph \textbf{T}ransformer (\textbf{SemHGT}), an LLM-enhanced framework for financial fraud detection over heterogeneous transaction graphs.

SemHGT follows three stages. First, role-aware prompts elicit risk-related business descriptions from node attributes, and a text encoder maps them to semantic vectors that enrich node representations. Second, semantic similarity guides multi-round sampling of target-specific path evidence under predefined transaction-to-transaction schemas. Third, for each schema, the sampled local evidence is combined with a global summary of the target's transaction neighbors to form a schema-level token. A unified transformer processes these tokens together with a token for the target transaction to predict fraud.

The main contributions can be summarized as follows:
\begin{itemize}
    \item We explore the use of LLM-derived, role-aware business semantics to enhance heterogeneous graph learning for financial fraud detection.
    \item We propose SemHGT, an LLM-enhanced heterogeneous graph transformer framework that makes role-aware business semantics explicit, using them to enrich node representations and guide target-conditioned path evidence selection for financial fraud detection.
    \item We evaluate SemHGT on three benchmark datasets, where it leads the evaluated baselines on most dataset--metric comparisons. Further analyses provide evidence for the effectiveness of the design and clarify how semantic modeling and evidence selection contribute to fraud detection.
\end{itemize}

\section{Related Work}
\label{sec:related-work}

\paperparagraph{GNNs for Financial Fraud Detection.}

Graph neural networks (GNNs) learn node representations by propagating and aggregating information over graph connections. Financial transaction graphs contain multiple entity and relation types, motivating heterogeneous graph neural networks (HGNNs) that preserve typed connectivity. Meta-path-based models aggregate information along predefined typed schemas: HAN~\cite{wang2019heterogeneous} combines meta-path neighbors and schemas through hierarchical attention, while MAGNN~\cite{fu2020magnn} encodes intermediate entities within each path instance. Meta-path-free models instead parameterize message passing using node and relation types, as in R-GCN~\cite{10.1007/978-3-319-93417-4_38}, HGT~\cite{hu2020heterogeneous}, and R-HGNN~\cite{9737399}. Further approaches, including SlotGAT~\cite{pmlr-v202-zhou23j}, HINormer~\cite{10.1145/3543507.3583493}, and PHGT~\cite{10.24963/ijcai.2024/247}, explore slot-based aggregation, sampled structural contexts, and schema-aware tokens. Together, these methods offer increasingly expressive ways to encode heterogeneous structure.

GNNs for financial fraud detection further adapt graph learning to fraud-specific challenges. GTAN~\cite{xiang2023semi} captures temporal transaction behavior with gated temporal attention; CARE-GNN~\cite{10.1145/3340531.3411903} uses label-aware neighbor selection to mitigate feature and relation camouflage; PC-GNN~\cite{10.1145/3442381.3449989} addresses class imbalance through label-balanced subgraph construction and sampling; and GAAP~\cite{duan2025global} models attribute-association patterns across graph neighborhoods. While these methods exploit temporal, neighborhood, and attribute-association signals, they do not explicitly incorporate semantic information as a complementary signal for fraud detection.

\paperparagraph{Integrating LLMs with Graphs.}

Research integrating large language models (LLMs) with graphs follows two complementary directions. Graph-enhanced LLMs incorporate graph structure or neighborhood information into language-model inputs or training. THLM~\cite{zou-etal-2023-pretraining} uses topology-aware context-graph prediction during pre-training, while HiGPT~\cite{10.1145/3637528.3671987} expresses heterogeneous graph tasks through graph instructions to support cross-graph transfer. For fraud detection, DGP~\cite{li2026dgp} uses dual-granularity prompts to retain detailed information about the target while summarizing its graph neighborhood, making graph context available to an LLM-based classifier. However, the computational cost of LLM-based inference can limit the scalability of graph-enhanced LLMs, particularly in large-scale industrial applications.

LLM-enhanced GNNs use language-model outputs to provide features, auxiliary supervision, or data for a graph encoder. TAPE~\cite{ICLR2024_1766d75b} converts LLM-generated explanations into features for GNNs on text-attributed graphs. Other approaches apply LLMs to heterogeneous graph learning by generating minority-class nodes and edges, as in LLM-HetGDT~\cite{ma-etal-2025-llm}, or completing missing attributes before graph encoding, as in HINAC~\cite{ou2026hinac}. MH-LGC~\cite{ou2026targeting} uses LLM-guided contrastive learning in a multi-view hypergraph detector. Together, these studies use LLMs to provide graph features, graph-conditioned predictions, training signals, or additional graph data. Despite this progress, most existing LLM-enhanced graph methods target homogeneous graphs or heterogeneous graphs whose nodes already carry natural-language text. Still, financial fraud detection often involves heterogeneous transaction graphs whose node attributes consist primarily of structured numerical and categorical fields rather than natural-language text. 
% Unlike homogeneous graphs with a single node type, heterogeneous transaction graphs preserve distinct entity and relation types that encode business roles and interactions. In this setting, it is important to use role-aware business semantics both to enrich node representations and to guide target-specific sampling of local evidence along typed paths. SemHGT addresses this need by combining these two uses of business semantics.

\section{Method}
\label{sec:method}

\subsection{Problem Formulation and Framework Overview}

Let $\mathcal{G}=(\mathcal{V},\mathcal{E},\mathcal{T},\mathcal{R})$ denote a heterogeneous transaction graph, where $\mathcal{V}$ and $\mathcal{E}$ are its node and edge sets, and $\mathcal{T}$ and $\mathcal{R}$ are its node- and relation-type sets. Each node $v\in\mathcal{V}$ has type $\tau(v)\in\mathcal{T}$ and structured attributes $\mathbf{x}_v$. Let $\mathcal{V}_{\mathrm{txn}}\subseteq\mathcal{V}$ be the transaction nodes. Given a target transaction $i\in\mathcal{V}_{\mathrm{txn}}$, the task is to predict its binary fraud label $y_i\in\{0,1\}$, where $1$ denotes fraud and $0$ denotes benign.

\begin{figure*}[t]
    \centering
    \includegraphics[width=\textwidth]{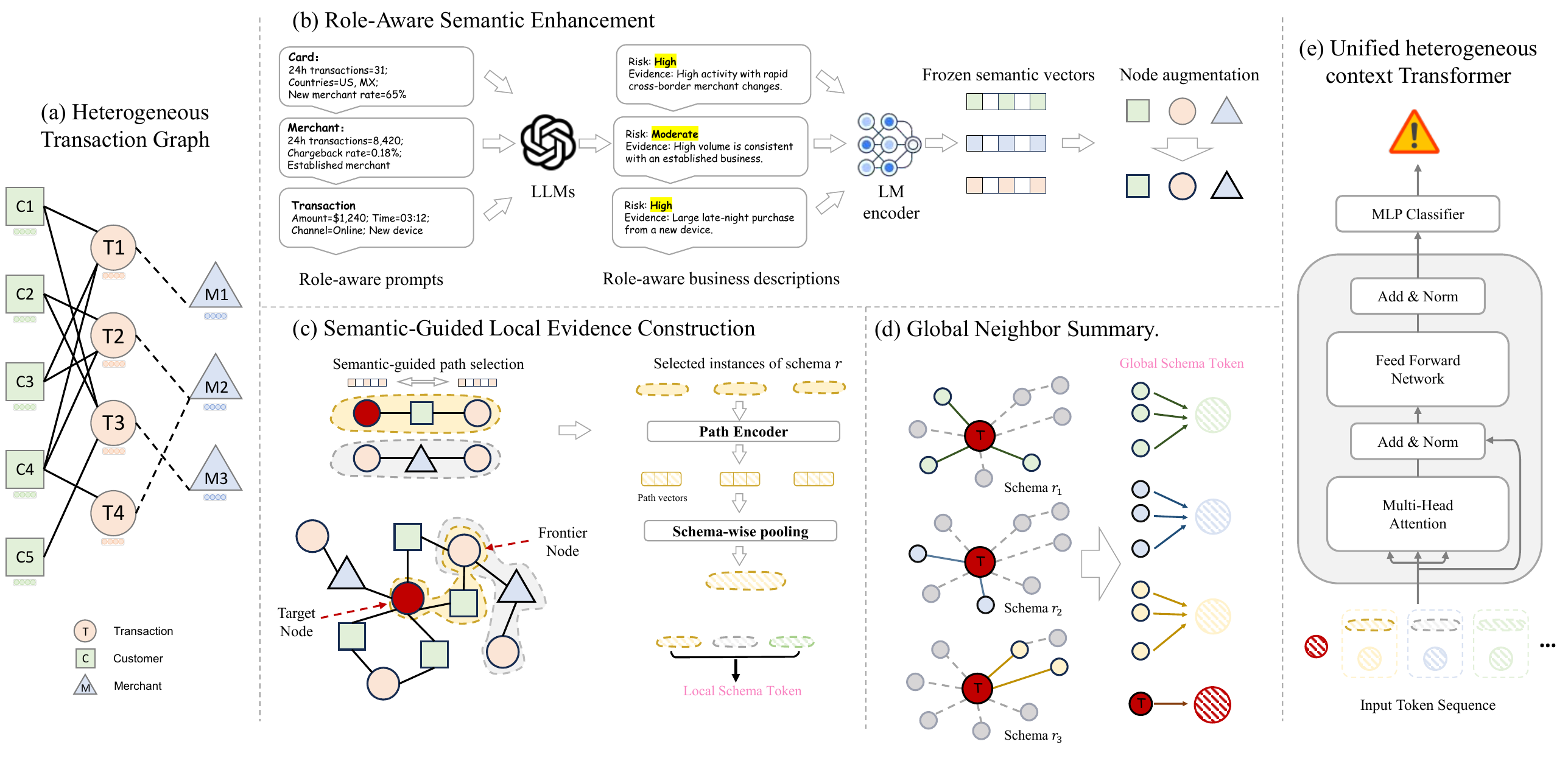}
    \caption{Overview of SemHGT. Role-aware business semantics enhance node representations and guide target-conditioned path sampling. For each schema, SemHGT fuses the sampled local path summary with the average representation of the target's transaction neighbors under that schema to construct a schema-level token. A unified transformer jointly processes the target token and all schema-level tokens for fraud prediction.}
    \Description{The SemHGT framework uses a language model and a frozen text encoder to derive role-aware business semantics from node attributes. These semantics enhance node features and guide path sampling. For each schema, the model fuses sampled local path evidence with the average representation of the target's transaction neighbors, then processes the resulting schema-level tokens with a target token using a unified transformer.}
    \label{fig:method}
\end{figure*}

Let $\mathcal{S}=\{r_1,\ldots,r_R\}$ be the predefined transaction-to-transaction path schemas. As shown in Figure~\ref{fig:method}, SemHGT follows three stages: it first enhances typed node representations with role-aware business semantics, then samples and encodes target-specific path evidence under each schema, and finally fuses each local path summary with the global summary of the target's transaction neighbors to form one schema-level token. A transformer processes these tokens together with a token representing the target transaction.

\begin{figure}[t]
    \centering
    \includegraphics[width=0.62\columnwidth]{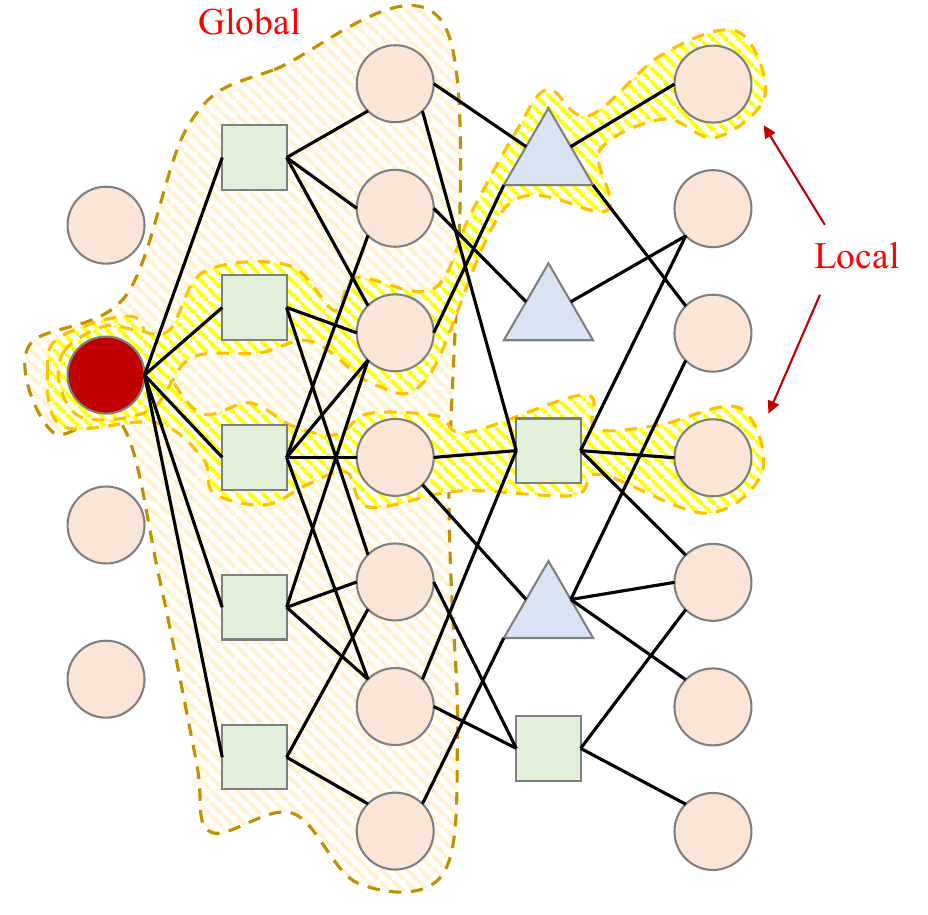}
    \caption{Global and local context for a target transaction. Global context averages the transaction neighbors under each schema, while local context captures the selected target-relevant path evidence.}
    \Description{A target transaction is shown with two complementary contexts: its transaction neighbors under each schema and selected local path evidence around the target.}
    \label{fig:global-local-context}
\end{figure}

\subsection{Role-Aware Semantic Enhancement}
\label{sec:semantic-node-enhancement}

\paperparagraph{LLM-Based Description Generation.}
The risk implications of an attribute depend on the entity's business role. We therefore use a role-aware prompt $\Phi_{\tau(v)}$ to interpret the attributes of node $v$ in light of corresponding reference statistics. The language model generates a risk-related business description, which a frozen text encoder maps to a normalized semantic vector:
\begin{equation}
    \mathbf{d}_v = \operatorname{LLM}\!\left(\Phi_{\tau(v)}(\mathbf{x}_v)\right),
    \qquad
    \mathbf{s}_v =
    \frac{\operatorname{Enc}(\mathbf{d}_v)}
    {\left\lVert\operatorname{Enc}(\mathbf{d}_v)\right\rVert_2}.
    \label{eq:semantic-embedding}
\end{equation}
The complete prompt templates are provided in Appendix~\ref{app:creditcard-prompt}.

\paperparagraph{Semantic Node Enhancement.}
For each node type, separate projections map the raw attributes and semantic vector to a common $d$-dimensional space, yielding raw-feature representation $\mathbf{r}_v$ and semantic representation $\mathbf{q}_v$:
\begin{equation}
    \mathbf{r}_v=f^{\mathrm{raw}}_{\tau(v)}(\mathbf{x}_v),
    \qquad
    \mathbf{q}_v=f^{\mathrm{sem}}_{\tau(v)}(\mathbf{s}_v).
    \label{eq:node-projections}
\end{equation}
We use a feature-wise gate to inject semantic information without discarding the structured representation:
\begin{equation}
    \mathbf{g}_v=\sigma\!\left(
        \mathbf{W}^{g}_{\tau(v)}[\mathbf{r}_v\Vert\mathbf{q}_v]
        +\mathbf{b}^{g}_{\tau(v)}
    \right),
    \qquad
    \mathbf{h}_v=\operatorname{LN}\!\left(
        \mathbf{r}_v+\alpha\,\mathbf{g}_v\odot\mathbf{q}_v
    \right).
    \label{eq:semantic-fusion}
\end{equation}
Here, $\mathbf{W}^{g}_{\tau(v)}$ and $\mathbf{b}^{g}_{\tau(v)}$ are type-specific learnable parameters, $\sigma$ is the sigmoid function, and $\alpha$ is a learnable scalar. The gate controls the semantic contribution to the raw-feature representation, producing the enhanced representation $\mathbf{h}_v$ used in path sampling and token construction.

\subsection{Semantic-Guided Local Evidence Construction}
\label{sec:path-evidence}

\paperparagraph{Semantic-Guided Path Sampling.}
For schema $r\in\mathcal{S}$, a path instance is an ordered typed sequence
\begin{equation}
    \xi=(u,a_1,\ldots,a_m,v), \qquad r(\xi)=r,
    \label{eq:path-instance}
\end{equation}
where $u,v\in\mathcal{V}_{\mathrm{txn}}$ are the endpoint transactions and $a_1,\ldots,a_m$ are the intermediate nodes along schema $r$. For example, $T$--$W$--$T$ denotes two transaction nodes ($T$) connected through a wallet node ($W$).

Starting from target $i$, the sampler expands for a fixed number of rounds. For a frontier transaction $u$ and a candidate endpoint transaction $v$, it scores the candidate by
\begin{equation}
    q_i(u,v)=
    \lambda\,
    \mathbf{s}_{u}^{\top}\mathbf{s}_{v}
    +
    (1-\lambda)\,
    \mathbf{s}_{i}^{\top}\mathbf{s}_{v}.
    \label{eq:routing-score}
\end{equation}
Here, $\lambda\in[0,1]$ controls the relative weight of similarity to the current frontier transaction $u$ and the target transaction $i$, with weights $\lambda$ and $1-\lambda$, respectively. Since the semantic vectors are normalized, both similarities are cosine similarities. Candidates are ranked independently per schema, and at most $k_r$ path instances are retained for schema $r$ in each sampling round, subject to its candidate cap and the shared context budget. The retained set for target $i$ and schema $r$, denoted by $\Xi_i^r$, accumulates the selected paths across rounds. Selected paths are added atomically, preserving their intermediate nodes under the node budget.

Each round expands from the transaction endpoints reached in the previous round, allowing the sampled evidence to span multiple relations, such as $T$--Card--$T$--Merchant--$T$.

\paperparagraph{Local Path Encoding and Pooling.}
To summarize the sampled local evidence, each retained instance $\xi=(u,a_1,\ldots,a_m,v)$ is encoded in typed order by a GRU~\cite{cho-etal-2014-learning}:
\begin{equation}
    \mathbf{u}_{i,\xi}=
    \operatorname{GRU}(\mathbf{h}_u,\mathbf{h}_{a_1},\ldots,\mathbf{h}_v)_{\mathrm{last}}.
    \label{eq:path-gru}
\end{equation}
The final GRU state $\mathbf{u}_{i,\xi}$ represents the path. For each schema, target-conditioned attention weights and pools the retained path states:

\begin{equation}
    a_{i,\xi}=\frac{(\mathbf{W}_p\mathbf{u}_{i,\xi})^{\top}
    (\mathbf{W}_h\mathbf{h}_i)}{\sqrt{d_a}},
    \qquad
    \omega_{i,\xi}=
    \operatorname{softmax}_{\xi\in\Xi_i^{r(\xi)}}(a_{i,\xi}),
    \label{eq:path-weight}
\end{equation}
\begin{equation}
    \mathbf{p}_{i,r}=
    \sum_{\xi\in\Xi_i^r}
    \omega_{i,\xi}\mathbf{u}_{i,\xi}.
    \label{eq:schema-pooling}
\end{equation}
Here, $\mathbf{W}_p$ and $\mathbf{W}_h$ project into a $d_a$-dimensional attention space, $a_{i,\xi}$ is the path--target score, $\omega_{i,\xi}$ is its within-schema normalized weight, and $\mathbf{p}_{i,r}$ is the pooled local summary for schema $r$.

\subsection{Schema-Token Construction and Prediction}
\label{sec:schema-token-construction}

\paperparagraph{Global Neighbor Summary.}
For each schema $r$, let $\mathcal{N}_r(i)$ contain the transaction neighbors of target $i$ in the observable graph induced by $r$. We summarize this neighborhood by averaging their enhanced representations:
\begin{equation}
    \mathbf{c}_{i,r}=
    \operatorname{Mean}\{\mathbf{h}_j:j\in\mathcal{N}_r(i)\},
    \qquad
    \mathbf{t}^{(0)}_{i,r}=f_{\mathrm{rel}}(\mathbf{c}_{i,r})+\mathbf{e}_r,
    \label{eq:relation-token}
\end{equation}
The learned projection $f_{\mathrm{rel}}$ maps the schema mean to the relation-token space, and the learned schema embedding $\mathbf{e}_r$ gives the initial token $\mathbf{t}^{(0)}_{i,r}$ its schema identity.

\paperparagraph{Local--Global Fusion.}
The pooled local path summary is fused only with the token of its own schema using a feature-wise gated residual:
\begin{equation}
    \boldsymbol{\gamma}_{i,r}=
    \sigma\!\left(
    \mathbf{W}_{\gamma}
    [\mathbf{t}^{(0)}_{i,r}\Vert\tilde{\mathbf{p}}_{i,r}]
    +\mathbf{b}_{\gamma}\right),
    \qquad
    \mathbf{t}_{i,r}=\mathbf{t}^{(0)}_{i,r}+
    \boldsymbol{\gamma}_{i,r}\odot\tilde{\mathbf{p}}_{i,r}.
    \label{eq:path-relation-fusion}
\end{equation}
Here, $\tilde{\mathbf{p}}_{i,r}=f_{\mathrm{adapt}}(\mathbf{p}_{i,r})$ is the projected local summary; $\mathbf{W}_{\gamma}$ and $\mathbf{b}_{\gamma}$ are learnable gate parameters; and $\boldsymbol{\gamma}_{i,r}$ controls its contribution. The resulting $\mathbf{t}_{i,r}$ is the final schema-level token.

\paperparagraph{Token Sequence and Prediction.}
We construct the root token as $\mathbf{t}_{i,0}=f_{\mathrm{root}}(\mathbf{h}_i)+\mathbf{e}_0$, where $f_{\mathrm{root}}$ is a learned projection and $\mathbf{e}_0$ is a learned root identity embedding. The initial transformer input contains this root token and one schema-level token for every enabled schema:
\begin{equation}
    \mathbf{X}^{(0)}_i=
    [\mathbf{t}_{i,0};\mathbf{t}_{i,r_1};\ldots;\mathbf{t}_{i,r_R}],
    \qquad \{r_1,\ldots,r_R\}=\mathcal{S}.
    \label{eq:transformer-input}
\end{equation}
A transformer encoder~\cite{NIPS2017_3f5ee243} processes the sequence, and the final state of the target token, $\mathbf{z}_i$, is used for classification:
\begin{equation}
    \boldsymbol{\ell}_i=f_{\mathrm{cls}}(\mathbf{z}_i),
    \qquad
    \widehat{p}_i=\operatorname{softmax}(\boldsymbol{\ell}_i)_1.
    \label{eq:prediction}
\end{equation}
The classifier $f_{\mathrm{cls}}$ maps the final root state to class logits $\boldsymbol{\ell}_i$; $\widehat{p}_i$ is the predicted fraud probability.

Algorithm~\ref{alg:semhgt} summarizes target-wise inference, from semantic enhancement and iterative path sampling to schema-token construction and fraud prediction.
\begin{algorithm}[t]
\caption{Target-wise Inference in SemHGT}
\label{alg:semhgt}
\begin{algorithmic}[1]
\Require Graph $\mathcal{G}$, target transaction $i$, schemas $\mathcal{S}$, and sampling rounds $L$
\Ensure $\widehat{p}_i$
\State Obtain enhanced node representations $\mathbf{h}_v$ (Eq.~\eqref{eq:semantic-fusion})
\State Initialize $\mathcal{F}_i\gets\{i\}$ and $\Xi_i^r\gets\emptyset$ for all $r\in\mathcal{S}$
\For{$\ell=1,\ldots,L$}
    \State Sample schema-consistent paths from $\mathcal{F}_i$ and score endpoints with $q_i(u,v)$ (Eq.~\eqref{eq:routing-score})
    \State Retain top-ranked paths under schema caps and context budget; update $\Xi_i^r$ and $\mathcal{F}_i$
\EndFor
\ForAll{$r\in\mathcal{S}$}
    \State Encode each retained path $\xi=(u,a_1,\ldots,v)$ with a GRU
    \Statex \hspace{\algorithmicindent} $\mathbf{u}_{i,\xi}\gets\operatorname{GRU}(\mathbf{h}_u,\mathbf{h}_{a_1},\ldots,\mathbf{h}_v)_{\mathrm{last}},\quad\xi\in\Xi_i^r$ (Eq.~\eqref{eq:path-gru})
    \State Pool path vectors with target-conditioned weights:
    \Statex \hspace{\algorithmicindent} $\mathbf{p}_{i,r}\gets\sum_{\xi\in\Xi_i^r}\omega_{i,\xi}\mathbf{u}_{i,\xi}$ (Eqs.~\eqref{eq:path-weight}--\eqref{eq:schema-pooling})
    \State Form an initial schema token from the global neighbor summary (Eq.~\eqref{eq:relation-token}) and fuse local evidence:
    \Statex \hspace{\algorithmicindent} $\mathbf{t}_{i,r}\gets\mathbf{t}^{(0)}_{i,r}+\boldsymbol{\gamma}_{i,r}\odot f_{\mathrm{adapt}}(\mathbf{p}_{i,r})$ (Eq.~\eqref{eq:path-relation-fusion})
\EndFor
\State Form $\mathbf{X}_i^{(0)}$ from the root and schema tokens (Eq.~\eqref{eq:transformer-input})
\State Obtain the final target-token representation:
\Statex \hspace{\algorithmicindent} $\mathbf{z}_i\gets[\operatorname{transformer}(\mathbf{X}_i^{(0)})]_0$
\State Compute $\widehat{p}_i\gets\operatorname{softmax}(f_{\mathrm{cls}}(\mathbf{z}_i))_1$
\State \Return $\widehat{p}_i$
\end{algorithmic}
\end{algorithm}

\paperparagraph{Training Objective.}

The node projections, semantic gates, path encoder, relation adapter, transformer, and classifier are trained jointly, while the language model and text encoder remain frozen. We optimize weighted cross-entropy over the labeled training transactions $\mathcal{V}_{\mathrm{train}}$:
\begin{equation}
    \mathcal{L}=-\frac{1}{|\mathcal{V}_{\mathrm{train}}|}
    \sum_{i\in\mathcal{V}_{\mathrm{train}}}w_{y_i}
    \log\!\left(\operatorname{softmax}(\boldsymbol{\ell}_i)_{y_i}\right),
    \label{eq:objective}
\end{equation}
where $w_{y_i}$ is the loss weight for class $y_i$.

\subsection{Target-wise Computational Complexity}
\label{sec:complexity}

For a target transaction $i$, let $C_i$ be the number of candidate paths scored across schemas and sampling rounds, and let $\Xi_i=\bigcup_{r\in\mathcal{S}}\Xi_i^r$ be the retained paths, with $P_i=|\Xi_i|$. Their average node count is $\bar{m}_i=P_i^{-1}\sum_{\xi\in\Xi_i}|\xi|$; for a path $\xi=(u,a_1,\ldots,a_m,v)$, $|\xi|=m+2$. Let $D_i=\sum_{r\in\mathcal{S}}|\mathcal{N}_r(i)|$ be the total number of schema-specific transaction neighbors used in the global summaries. Scoring candidates costs $O(C_i d_s)$, encoding the retained paths with the GRU costs $O(P_i\bar{m}_i d^2)$, and aggregating global neighbors costs $O(D_i d)$. Thus, for fixed sampling rounds, schemas, and model dimensions, the leading target-dependent computation is
\begin{equation}
    T_i=O\!\left(C_i d_s+P_i\bar{m}_i d^2+D_i d\right),
    \label{eq:complexity}
\end{equation}
where $d_s$ is the semantic-vector dimension and $d$ is the hidden dimension. The transformer processes a fixed-length sequence of the target token and schema tokens, so its cost does not grow with the number of sampled paths.

\section{Experiments}
\label{sec:experiments}

\subsection{Experimental Setup}
\label{sec:experimental-setup}

\paperparagraph{Datasets.}
We evaluate SemHGT on CreditCard, BankTrans, and Elliptic++. CreditCard and BankTrans are the heterogeneous financial fraud datasets constructed in FinFraudBench~\cite{chen2026finfraudbench}, while Elliptic++ is a transaction-and-wallet benchmark~\cite{10.1145/3580305.3599803}. Each task is binary classification over transaction nodes. CreditCard and Elliptic++ are split chronologically into training, validation, and test periods in an approximate 6:1:3 ratio under an inductive protocol~\cite{NIPS2017_5dd9db5e}, whereas BankTrans uses a label-stratified 1:1:8 split under a transductive protocol~\cite{kipf2016semi}. In BankTrans, validation and test features and topology are available as unlabeled context during training, while their labels remain withheld. CreditCard and Elliptic++ are evaluated inductively on separate graph components for each temporal split, with no cross-split edges; both use the same split-aware policy for label-blind reference statistics. Table~\ref{tab:graph-characteristics} summarizes graph sizes, transaction-label composition, and metapath schemas; further graph and split details appear in Appendix~\ref{app:experimental-details}.

\begin{table}[t]
  \centering
  \caption{Statistics of the experimental graph views.}
  \label{tab:graph-characteristics}

  \scriptsize
  \setlength{\tabcolsep}{2pt}
  \renewcommand{\arraystretch}{1.12}

  % 按节点信息中最长一行确定三个文字块的统一宽度
  \settowidth{\nodecontentwidth}{Transaction: 1,852,394}

  \begin{tabular}{
    @{}
    >{\raggedright\arraybackslash}m{40pt}
    >{\centering\arraybackslash}m{75pt}
    >{\centering\arraybackslash}m{48pt}
    >{\centering\arraybackslash}m{55pt}
    @{}
  }
    \toprule
    Dataset & Nodes & Edges & Metapaths \\
    \midrule

    CreditCard
    & \makebox[\nodecontentwidth][l]{%
        \shortstack[l]{
          \# node types: 6\\
          \# nodes: 1,860,300\\
          Transaction: 1,852,394\\
          Fraud: 9,651 (0.52\%)\\
          Customer: 2,817\\
          Card: 2,817\\
          Merchant: 2,079\\
          Category: 42\\
          Location: 151
        }%
      }
    & \shortstack[l]{
        \# edge types: 14\\
        \# edges:\\
        18,535,208
      }
    & \shortstack[c]{
        $T$--Customer--$T$\\
        $T$--Card--$T$\\
        $T$--Merchant--$T$\\
        $T$--Category--$T$\\
        $T$--Location--$T$
      } \\

    \midrule

    BankTrans
    & \makebox[\nodecontentwidth][l]{%
        \shortstack[l]{
          \# node types: 6\\
          \# nodes: 8,987,098\\
          Transaction: 8,914,963\\
          Fraud: 13,332 (0.15\%)\\
          Customer: 1,219\\
          Card: 4,070\\
          Merchant: 66,537\\
          Category: 109\\
          Location: 200
        }%
      }
    & \shortstack[l]{
        \# edge types: 14\\
        \# edges:\\
        89,227,542
      }
    & \shortstack[c]{
        $T$--Customer--$T$\\
        $T$--Card--$T$\\
        $T$--Merchant--$T$\\
        $T$--Category--$T$\\
        $T$--Location--$T$
      } \\

    \midrule

    Elliptic++
    & \makebox[\nodecontentwidth][l]{%
        \shortstack[l]{
          \# node types: 2\\
          \# nodes: 110,857\\
          Transaction: 46,564\\
          Fraud: 4,545 (9.76\%)\\
          Wallet: 64,293
        }%
      }
    & \shortstack[l]{
        \# edge types: 8\\
        \# edges:\\
        402,882
      }
    & \shortstack[c]{
        $T$--$T$\\
        $T$--$W$--$T$\\
        $T$--$W$--$W$--$T$\\
        $T$--$W$--$W$--$W$--$T$
      } \\

    \bottomrule
  \end{tabular}
\end{table}

The main results use all transactions in each test split. The ablation study also uses the full CreditCard test split; all other analytical experiments use fixed, class-balanced subsets of their corresponding test sets. Appendix~\ref{app:analysis-subsets} reports the subset sizes and class counts.

\paperparagraph{Evaluation Metrics.}
We report Macro-F1, Fraud-F1, area under the precision--recall curve (AUPRC), and area under the receiver operating characteristic curve (AUROC). Macro-F1 averages the F1 scores of the benign and fraud classes equally, while Fraud-F1 focuses on precision and recall for the fraud class at the validation-selected threshold. AUPRC summarizes the precision--recall trade-off across thresholds and is informative under class imbalance~\cite{10.1145/1143844.1143874}. AUROC measures how well the model ranks fraudulent transactions above benign ones across thresholds, complementing the threshold-dependent F1 measures.

\paperparagraph{Implementation Details.}
We use Qwen3-8B~\cite{yang2025qwen3} to generate risk-related business descriptions and a frozen MPNet encoder~\cite{NEURIPS2020_c3a690be} to map them to semantic vectors. The encoder produces 768-dimensional vectors, which are projected to 128 dimensions. The prediction transformer has two layers, four attention heads, a 128-dimensional hidden state, a feed-forward width of 256, and dropout 0.2. We train all datasets with AdamW~\cite{loshchilov2017decoupled} and weight decay $10^{-4}$. Table~\ref{tab:training-budgets} summarizes the epoch budgets, batch sizes, and learning rates.

\begin{table}[!b]
\centering
\caption{Core dataset-specific training settings; early-stopping patience is shown in parentheses.}
\label{tab:training-budgets}
\scriptsize
\setlength{\tabcolsep}{3pt}
\begin{tabular}{@{}lrrl@{}}
\toprule
Dataset & Max epochs (patience) & Batch size & Learning rate \\
\midrule
CreditCard & 40 (10) & 256 & $10^{-3}$ (constant) \\
BankTrans & 50 (10) & 128 & $10^{-4}\!\to\!10^{-5}$ (cosine) \\
Elliptic++ & 40 (10) & 128 & $3\times10^{-4}$ (constant) \\
\bottomrule
\end{tabular}
\end{table}

The local path schemas are dataset-specific. CreditCard and BankTrans use five schemas: $T$--Customer--$T$ (same customer), $T$--Card--$T$ (same card), $T$--Merchant--$T$ (same merchant), $T$--Category--$T$ (same category), and $T$--Location--$T$ (same location). Elliptic++ uses four schemas: $T$--$T$ (direct transaction link), $T$--$W$--$T$ (transactions sharing a wallet), $T$--$W$--$W$--$T$ (transactions connected through two intermediate wallets), and $T$--$W$--$W$--$W$--$T$ (transactions connected through three intermediate wallets). For each schema, the global component averages the representations of its transaction neighbors; the sampled local path summary is then fused into the corresponding schema-level token.

\begin{table*}[!t]
\centering
\caption{Main fraud-detection results on CreditCard, BankTrans, and Elliptic++. For each dataset, the table reports Macro-F1, Fraud-F1, AUPRC, and AUROC. All values are percentages, shown as mean $\pm$ standard deviation across three runs. The best result in each metric column is boldfaced and shaded, and the second-best is underlined.}
\label{tab:main-results}
\footnotesize
\setlength{\tabcolsep}{2pt}
\renewcommand{\arraystretch}{1.12}
\resizebox{\textwidth}{!}{%
\begin{tabular}{lrrrrrrrrrrrr}
\toprule
\multirow{2}{*}{Method} & \multicolumn{4}{c}{CreditCard} & \multicolumn{4}{c}{BankTrans} & \multicolumn{4}{c}{Elliptic++} \\
\cmidrule(lr){2-5}\cmidrule(lr){6-9}\cmidrule(lr){10-13}
 & Macro-F1 & Fraud-F1 & AUPRC & AUROC & Macro-F1 & Fraud-F1 & AUPRC & AUROC & Macro-F1 & Fraud-F1 & AUPRC & AUROC \\
\midrule
MLP & \ms{72.41}{0.15} & \ms{44.99}{0.31} & \ms{40.83}{0.24} & \ms{94.29}{0.14} & \ms{54.47}{0.67} & \ms{9.17}{1.41} & \ms{3.76}{0.02} & \ms{92.75}{0.02} & \ms{79.10}{0.61} & \ms{60.31}{1.20} & \ms{59.57}{0.59} & \ms{84.49}{0.41} \\
SlotGAT~\cite{pmlr-v202-zhou23j} & \ms{76.62}{0.47} & \ms{53.43}{0.93} & \ms{47.68}{1.01} & \ms{94.46}{0.51} & \ms{86.64}{0.61} & \ms{73.32}{1.21} & \ms{67.95}{1.83} & \ms{96.82}{0.11} & \ms{74.25}{3.15} & \ms{51.08}{6.06} & \ms{49.82}{7.35} & \ms{86.20}{2.23} \\
SimpleHGN~\cite{10.1145/3447548.3467350} & \ms{63.31}{0.89} & \ms{26.89}{1.78} & \ms{14.75}{0.40} & \ms{84.80}{0.14} & \ms{83.93}{0.65} & \ms{67.92}{1.29} & \ms{55.35}{6.37} & \ms{96.02}{0.20} & \ms{74.29}{0.49} & \ms{51.55}{0.91} & \ms{52.30}{0.74} & \ms{87.56}{0.35} \\
FastGTN~\cite{YUN2022104} & \ms{73.91}{0.34} & \ms{48.02}{0.68} & \ms{47.72}{0.70} & \ms{93.76}{0.12} & \ms{57.31}{1.59} & \ms{14.84}{3.06} & \ms{6.45}{1.59} & \ms{94.76}{0.40} & \ms{77.23}{0.41} & \ms{56.56}{0.80} & \ms{60.20}{0.77} & \ms{84.99}{0.34} \\
HAN~\cite{wang2019heterogeneous} & \underline{\ms{83.34}{0.88}} & \underline{\ms{66.81}{1.75}} & \underline{\ms{70.29}{1.72}} & \ms{96.97}{0.34} & \ms{80.25}{0.42} & \ms{60.55}{0.84} & \ms{54.74}{2.47} & \ms{94.35}{0.41} & \ms{71.71}{0.22} & \ms{47.58}{0.42} & \ms{55.34}{1.37} & \ms{85.30}{0.21} \\
SeHGNN~\cite{yang2023simple} & \ms{81.36}{0.75} & \ms{62.86}{1.50} & \ms{63.73}{1.41} & \ms{97.04}{0.10} & \ms{82.31}{0.74} & \ms{64.67}{1.47} & \ms{63.97}{0.53} & \ms{95.86}{0.25} & \ms{74.99}{0.66} & \ms{52.83}{1.17} & \ms{57.94}{1.89} & \ms{86.52}{0.65} \\
R-GCN~\cite{10.1007/978-3-319-93417-4_38} & \ms{77.11}{1.40} & \ms{54.40}{2.79} & \ms{55.28}{2.28} & \ms{95.65}{0.56} & \ms{87.30}{0.25} & \ms{74.64}{0.50} & \ms{69.95}{2.00} & \ms{96.54}{0.10} & \ms{61.86}{0.76} & \ms{33.72}{0.98} & \ms{59.87}{1.80} & \ms{85.74}{0.81} \\
GCN~\cite{kipf2016semi} & \ms{80.34}{0.20} & \ms{60.80}{0.41} & \ms{59.55}{0.90} & \ms{95.03}{0.15} & \ms{68.33}{0.77} & \ms{36.75}{1.53} & \ms{30.41}{2.28} & \ms{96.11}{0.06} & \ms{77.09}{0.08} & \ms{56.51}{0.09} & \ms{58.51}{0.84} & \ms{85.99}{0.71} \\
R-HGNN~\cite{9737399} & \ms{79.37}{0.23} & \ms{58.90}{0.45} & \ms{59.09}{0.48} & \ms{95.12}{0.82} & \ms{86.12}{1.40} & \ms{72.29}{2.80} & \ms{68.14}{6.37} & \ms{96.90}{0.04} & \ms{74.08}{0.92} & \ms{50.97}{1.71} & \ms{52.68}{1.54} & \ms{87.20}{0.49} \\
HeSRN~\cite{lu2025hesrn} & \ms{52.62}{3.16} & \ms{8.02}{3.97} & \ms{14.78}{8.17} & \ms{88.61}{5.13} & \ms{82.38}{5.40} & \ms{64.81}{10.79} & \ms{59.03}{12.61} & \ms{95.67}{0.86} & \ms{70.30}{0.21} & \ms{44.73}{0.43} & \ms{43.04}{3.08} & \ms{86.26}{0.33} \\
NARS~\cite{yu2020scalable} & \ms{78.73}{0.27} & \ms{57.61}{0.52} & \ms{59.10}{0.19} & \ms{93.33}{0.42} & \ms{75.58}{0.52} & \ms{51.23}{1.04} & \ms{44.06}{1.57} & \ms{97.00}{0.15} & \underline{\ms{80.11}{0.28}} & \underline{\ms{62.30}{0.50}} & \ms{62.15}{0.77} & \ms{86.26}{0.20} \\
LSPI~\cite{zhao2025lspi} & \ms{56.05}{1.19} & \ms{12.29}{2.39} & \ms{25.25}{24.94} & \ms{72.51}{25.30} & \ms{87.50}{0.35} & \ms{75.03}{0.70} & \underline{\ms{74.71}{0.60}} & \ms{96.90}{0.44} & \ms{70.77}{1.39} & \ms{44.82}{2.84} & \ms{48.60}{1.96} & \ms{86.90}{0.07} \\
HetTree~\cite{guan2025heterogeneous} & \ms{63.11}{1.25} & \ms{26.38}{2.49} & \ms{52.46}{3.41} & \ms{92.14}{0.99} & \ms{85.31}{0.84} & \ms{70.65}{1.67} & \ms{65.09}{5.40} & \ms{96.79}{0.22} & \ms{74.45}{0.46} & \ms{51.62}{0.85} & \ms{49.73}{1.88} & \ms{85.84}{0.11} \\
MLED~\cite{10.1145/3746027.3755245} & \ms{65.79}{0.42} & \ms{31.86}{0.84} & \ms{14.25}{0.96} & \ms{84.72}{0.36} & \ms{77.72}{2.12} & \ms{55.49}{4.24} & \ms{54.13}{6.19} & \ms{96.40}{0.37} & \ms{72.71}{1.11} & \ms{49.51}{1.85} & \ms{59.06}{0.97} & \ms{86.73}{0.02} \\
HetGNN~\cite{10.1145/3292500.3330961} & \ms{66.90}{6.01} & \ms{34.29}{11.72} & \ms{44.19}{2.28} & \ms{96.10}{2.34} & \ms{86.73}{0.45} & \ms{73.50}{0.89} & \ms{61.76}{0.03} & \ms{96.87}{0.25} & \ms{75.19}{0.68} & \ms{53.29}{1.33} & \ms{52.15}{1.95} & \ms{87.13}{0.35} \\
GTN~\cite{NEURIPS2019_9d63484a} & \ms{68.90}{8.26} & \ms{37.96}{16.50} & \ms{52.59}{3.37} & \ms{95.50}{1.11} & \underline{\ms{87.81}{0.57}} & \underline{\ms{75.65}{1.14}} & \ms{73.45}{1.73} & \best{\ms{97.37}{0.21}} & \ms{76.88}{1.95} & \ms{55.94}{3.75} & \ms{58.52}{3.71} & \underline{\ms{88.73}{0.98}} \\
HGT~\cite{hu2020heterogeneous} & \ms{65.49}{0.93} & \ms{31.30}{1.85} & \ms{13.79}{1.32} & \ms{84.18}{0.75} & \ms{78.45}{1.79} & \ms{56.95}{3.57} & \ms{56.01}{1.44} & \underline{\ms{97.30}{1.10}} & \ms{76.56}{0.08} & \ms{55.64}{0.11} & \ms{55.96}{1.94} & \ms{86.34}{0.87} \\
GAAP~\cite{duan2025global} & \ms{80.40}{0.50} & \ms{60.93}{0.99} & \ms{62.04}{1.01} & \underline{\ms{97.82}{0.14}} & \ms{73.85}{1.84} & \ms{47.77}{3.67} & \ms{39.38}{4.48} & \ms{96.83}{0.22} & \ms{78.11}{0.73} & \ms{58.24}{1.42} & \underline{\ms{62.84}{0.32}} & \ms{84.96}{0.44} \\
\midrule
SemHGT & \best{\ms{85.89}{0.70}} & \best{\ms{71.89}{1.38}} & \best{\ms{77.53}{0.98}} & \best{\ms{98.46}{0.20}} & \best{\ms{90.35}{0.09}} & \best{\ms{80.72}{0.19}} & \best{\ms{78.08}{0.24}} & \ms{96.86}{0.43} & \best{\ms{83.49}{0.27}} & \best{\ms{68.77}{0.52}} & \best{\ms{69.23}{1.14}} & \best{\ms{90.00}{0.72}} \\
\bottomrule
\end{tabular}%
}
\end{table*}
\paperparagraph{Baselines.}
The 18 baselines in Table~\ref{tab:main-results} cover several modeling paradigms. MLP uses transaction attributes without graph propagation, and GCN~\cite{kipf2016semi} represents homogeneous graph learning. The heterogeneous graph baselines include relation- and type-aware models such as R-GCN, HAN, GTN, FastGTN, HGT, HetGNN, SimpleHGN, SeHGNN, R-HGNN, and NARS~\cite{10.1007/978-3-319-93417-4_38,wang2019heterogeneous,NEURIPS2019_9d63484a,YUN2022104,hu2020heterogeneous,10.1145/3292500.3330961,10.1145/3447548.3467350,yang2023simple,9737399,yu2020scalable}. We also compare with fraud-oriented graph methods, including SlotGAT, HeSRN, LSPI, HetTree, MLED, and GAAP~\cite{pmlr-v202-zhou23j,lu2025hesrn,zhao2025lspi,guan2025heterogeneous,10.1145/3746027.3755245,duan2025global}.

We train and evaluate the models using three random seeds. The language model and text encoder remain frozen. Training labels alone are used for optimization. We select checkpoints by corrected validation AUPRC and calibrate the F1 threshold on validation data; final metrics are computed over every test transaction, with test labels used only for this evaluation. We report the mean and sample standard deviation across seeds. Appendix~\ref{app:experimental-details} provides the per-schema sampling budgets, validation protocol, and model-training configuration.

\subsection{Main Results}
\label{sec:main-results}

Table~\ref{tab:main-results} compares SemHGT with the evaluated baselines on CreditCard, BankTrans, and Elliptic++. SemHGT leads all four metrics on CreditCard and Elliptic++. On BankTrans, it achieves the strongest Macro-F1, Fraud-F1, and AUPRC, while GTN attains a higher AUROC (97.37 versus 96.86).

The margins over the strongest baseline further show gains across multiple aspects of detection. On CreditCard, SemHGT improves Macro-F1, Fraud-F1, AUPRC, and AUROC by 2.55, 5.08, 7.24, and 0.64 percentage points, respectively. The corresponding gains on BankTrans are 2.54, 5.07, and 3.37 points for Macro-F1, Fraud-F1, and AUPRC; on Elliptic++, they are 3.38, 6.47, 6.39, and 1.27 points across the four metrics. The consistent gains in Fraud-F1 and AUPRC provide complementary evidence: Fraud-F1 reflects the precision–recall balance for fraudulent transactions at the validation-calibrated threshold, while AUPRC summarizes this trade-off across thresholds. Their joint improvement therefore indicates gains both at the selected operating point and in ranking fraud cases, which is particularly relevant under class imbalance. Taken together, these results support the effectiveness of SemHGT for heterogeneous financial fraud detection.

\FloatBarrier
\begin{figure}[bt]
    \centering
    \includegraphics[width=0.90\columnwidth]{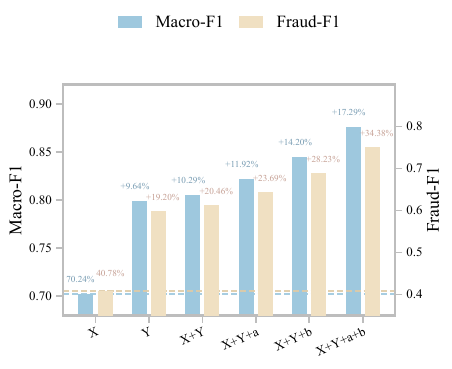}
    \caption{Ablation results on the full CreditCard test split for Macro-F1 and Fraud-F1 across six configurations combining local path evidence ($X$), global neighbor summaries ($Y$), and semantic enhancements. Dashed lines mark the $X$-only baseline, and annotations report the absolute percentage-point changes from this baseline for each configuration.}
    \Description{Grouped bars compare Macro-F1 and Fraud-F1 across X, Y, X plus Y, X plus Y plus node-semantic enhancement, X plus Y plus semantic-guided local path sampling, and the full setting. X denotes local path evidence and Y global neighbor summaries. The full setting performs best on both metrics, and dashed lines mark the X-only baseline.}
    \label{fig:ablation}
\end{figure}

\begin{figure*}[!t]
    \centering
    \includegraphics[width=1.0\textwidth]{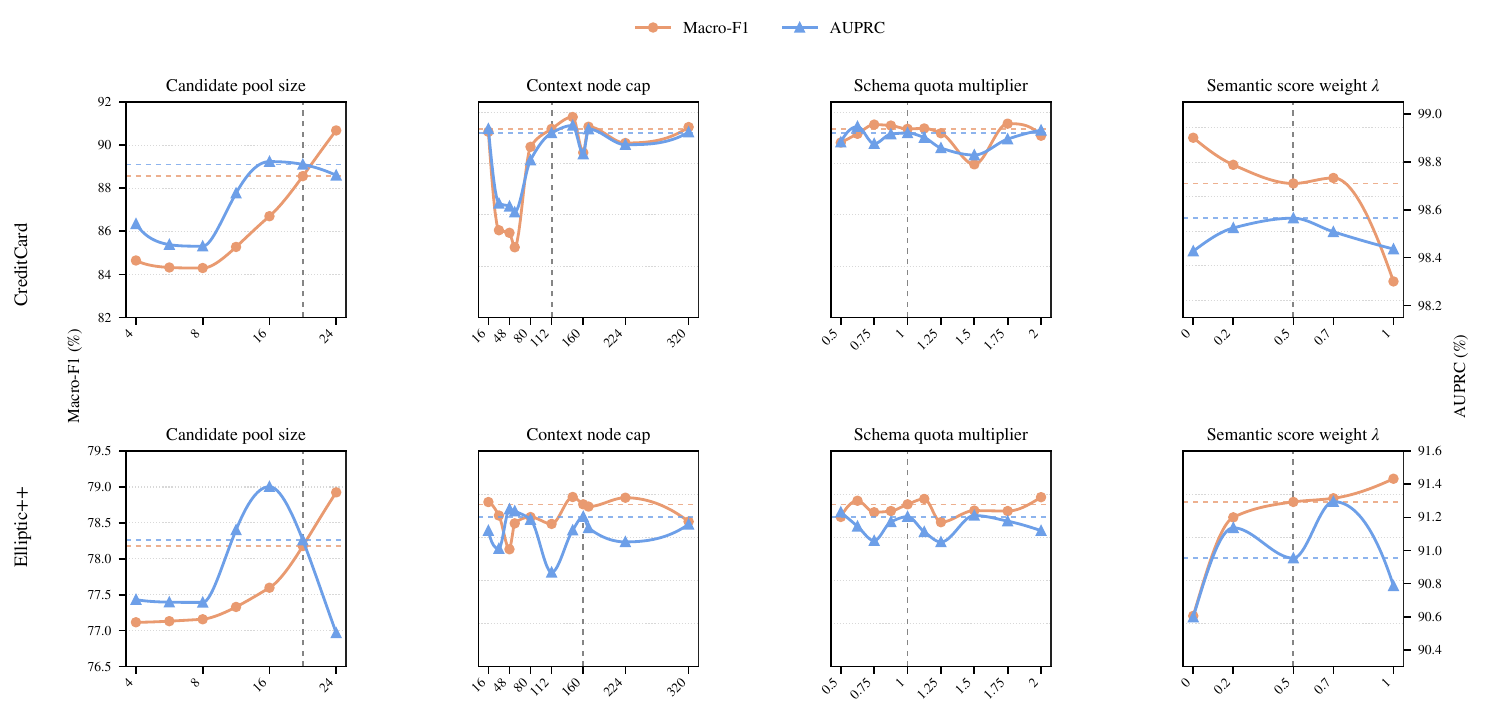}
    \caption{Hyperparameter sensitivity on the fixed, class-balanced test subsets of CreditCard (top) and Elliptic++ (bottom). From left to right, the columns vary candidate-pool size, context-node cap, schema-quota multiplier, and semantic-score weight $\lambda$. Orange circles show Macro-F1 (left axis), and blue triangles show AUPRC (right axis). Vertical dashed lines mark the reference setting for each parameter, while horizontal dashed lines indicate the corresponding metric scores.}
    \Description{A two-row, four-column grid shows Macro-F1 and AUPRC versus candidate pool size, context node cap, schema quota multiplier, and semantic score weight for CreditCard and Elliptic++. Candidate-pool Macro-F1 generally rises over the displayed range, context and quota responses are nonmonotonic, and intermediate semantic weights yield higher AUPRC than endpoints.}
    \label{fig:hyperparameter-sensitivity}
\end{figure*}

\subsection{Ablation Study}
\label{sec:ablation}

We run six controlled ablations on the full CreditCard test split to isolate the local and global components of the schema tokens and the two semantic enhancements. Let $X$ denote schema-level tokens built solely from sampled local path evidence, and $Y$ denote schema-level tokens built from the average representations of transaction neighbors under each schema. We compare $X$, $Y$, $X+Y$, $X+Y+a$, $X+Y+b$, and the full configuration $X+Y+a+b$, where $a$ enables business-semantic node enhancement and $b$ enables semantic-similarity guidance during local path sampling. When $b$ is disabled, paths are sampled uniformly with the same schemas and sampling budget. Figure~\ref{fig:ablation} reports Macro-F1 and Fraud-F1 for all six settings, enabling comparison of component effects.

$Y$ alone outperforms $X$ on both metrics, and combining the global and local components improves further. Adding either node-semantic enhancement or semantic-guided local path sampling to $X+Y$ yields additional gains, with the latter producing the larger improvement of the two. The full configuration achieves the highest scores, 87.53 Macro-F1 and 75.16 Fraud-F1, improving over $X$ by 17.29 and 34.38 percentage points, respectively.

$Y$ summarizes schema-specific transaction-neighbor context, while $X$ retains selected typed-path evidence. The semantic modules act at different stages: $a$ enriches node representations, whereas $b$ guides target-specific path selection. The full configuration's best scores support combining both semantic enhancements with local and global relational evidence.

\subsection{Hyperparameter Sensitivity}
\label{sec:hyperparameter-sensitivity}

We study four sampling parameters on the fixed, class-balanced CreditCard and Elliptic++ test subsets described above. The candidate-pool size controls how many path candidates are considered per schema, the context-node cap limits the total nodes retained in the sampled context, and the schema-quota multiplier scales the path quota for each schema. The semantic-score weight $\lambda$ balances candidate similarity to the current frontier and to the target transaction. Figure~\ref{fig:hyperparameter-sensitivity} reports Macro-F1 and AUPRC for each setting.

Increasing the candidate-pool size from 4 to 24 produces an overall rise in Macro-F1 on both datasets, although the intermediate settings are not strictly monotonic. AUPRC responds differently: it changes only modestly on CreditCard and peaks at an intermediate pool size on Elliptic++. Thus, a broader candidate pool can improve classification performance, but its effect is not uniform across ranking metrics.

The context-node cap and schema-quota multiplier also yield non-monotonic responses. Increasing either budget does not consistently improve Macro-F1 or AUPRC, and the more favorable settings differ across datasets and metrics. A possible explanation is a coverage--relevance trade-off: relaxing these constraints expands the available local evidence, but the additional nodes or paths may be less informative and introduce noise. Sampling performance therefore depends not only on the total evidence budget, but also on its allocation across schemas for each target transaction.

The semantic-score weight $\lambda$ balances similarity to the current frontier and the target transaction in Eq.~\eqref{eq:routing-score}. On the plotted sweep, the intermediate settings ($\lambda\in\{0.2,0.5,0.7\}$) yield higher AUPRC than both endpoints ($\lambda=0$ and $1$) on both datasets. At $\lambda=0.5$, for example, AUPRC reaches 98.565\% on CreditCard and 90.954\% on Elliptic++, compared with 98.427\%/98.436\% and 90.598\%/90.787\% at $\lambda=0/1$, respectively. This pattern supports combining target and frontier semantic similarities rather than relying exclusively on either. Macro-F1 is less consistent: among the plotted settings, $\lambda=0$ performs best on CreditCard and $\lambda=1$ on Elliptic++, so the advantage of intermediate weights is clearest for AUPRC.

\subsection{Sensitivity to Semantic Misalignment}
\label{sec:semantic-mismatch}

To measure sensitivity to node--semantic alignment, we interpolate each node's semantic vector with that of a randomly permuted node of the same type. For node $v$, the perturbed vector is
\begin{equation}
    \widetilde{\mathbf{s}}_v(\delta)
    =(1-\delta)\mathbf{s}_v
    +\delta\mathbf{s}_{\pi_{\tau(v)}(v)},
    \qquad \delta\in[0,1],
    \label{eq:semantic-mismatch}
\end{equation}
where $\pi_{\tau(v)}$ denotes a permutation within node type. The same permutation is used for all values of $\delta$ and is independent of fraud labels. For each setting, we train and evaluate the model using the perturbed vectors for both semantic node enhancement and semantic-guided path sampling; checkpoint selection and threshold calibration follow the validation protocol. The analysis uses the fixed, class-balanced CreditCard and Elliptic++ test subsets.

\begin{figure}[tb]
    \centering
    \includegraphics[width=0.80\columnwidth]{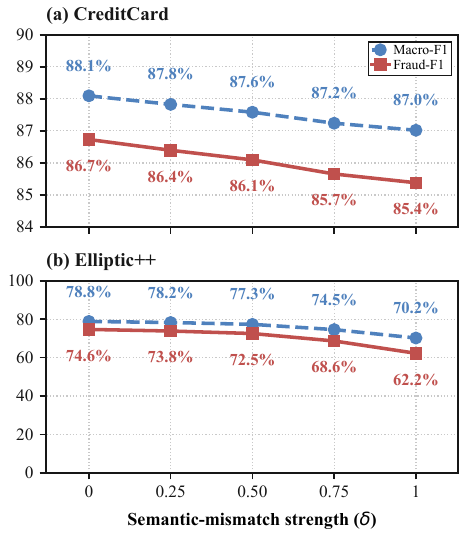}
    \caption{Sensitivity to within-type semantic mismatch on the balanced CreditCard and Elliptic++ test subsets. The mismatch strength $\delta$ controls the interpolation between each node's semantic vector and that of a randomly permuted node of the same type. Lines show Macro-F1 and Fraud-F1 as semantic alignment decreases.}
    \Description{Two vertically arranged line charts show Macro-F1 and Fraud-F1 against semantic mismatch strength $\delta$ for CreditCard (top) and Elliptic++ (bottom). Both metrics decline as mismatch strength increases.}
    \label{fig:semantic-mismatch}
\end{figure}

Figure~\ref{fig:semantic-mismatch} and Table~\ref{tab:semantic-mismatch-results} show that both metrics decline as $\delta$ increases, indicating that performance benefits from preserving the alignment between each node and its own risk-related business semantics. The decrease is modest on CreditCard and more pronounced on Elliptic++, where Macro-F1 falls from 78.80\% to 70.17\% and Fraud-F1 from 74.64\% to 62.22\% between the two endpoints.

\begin{table}[tb]
    \centering
    \caption{Macro-F1 and Fraud-F1 scores (\%) under within-type semantic mismatch on the balanced CreditCard and Elliptic++ test subsets. $\delta$ interpolates each node's semantic vector with a same-type permuted vector, from original alignment ($\delta=0$) to the permuted vector ($\delta=1$).}
    \label{tab:semantic-mismatch-results}
    \footnotesize
    \setlength{\tabcolsep}{4pt}
    \begin{tabular}{llrrrrr}
        \toprule
        Dataset & Metric & $\delta=0$ & $\delta=0.25$ & $\delta=0.50$ & $\delta=0.75$ & $\delta=1$ \\
        \midrule
        CreditCard & Macro-F1 & 88.10 & 87.83 & 87.58 & 87.24 & 87.02 \\
        CreditCard & Fraud-F1 & 86.73 & 86.39 & 86.09 & 85.65 & 85.38 \\
        Elliptic++ & Macro-F1 & 78.80 & 78.24 & 77.28 & 74.54 & 70.17 \\
        Elliptic++ & Fraud-F1 & 74.64 & 73.81 & 72.50 & 68.62 & 62.22 \\
        \bottomrule
    \end{tabular}
\end{table}

\subsection{Fraud-Label Composition of Sampled Evidence}
\label{sec:endpoint-composition}

We examine whether semantic-guided sampling changes the fraud-label composition of local evidence. For each target and schema, we measure the fraction of fraudulent terminal transaction nodes among the retained paths, excluding the target itself and duplicate endpoints within each sampling repeat. Semantic-guided and uniform sampling use the same schemas, candidate pools, path quotas, and context budgets; labels are used only to form the balanced target subsets and to analyze endpoints after sampling, never to compute sampling scores.

\begin{figure}[t]
    \centering
    \includegraphics[width=1.0\columnwidth]{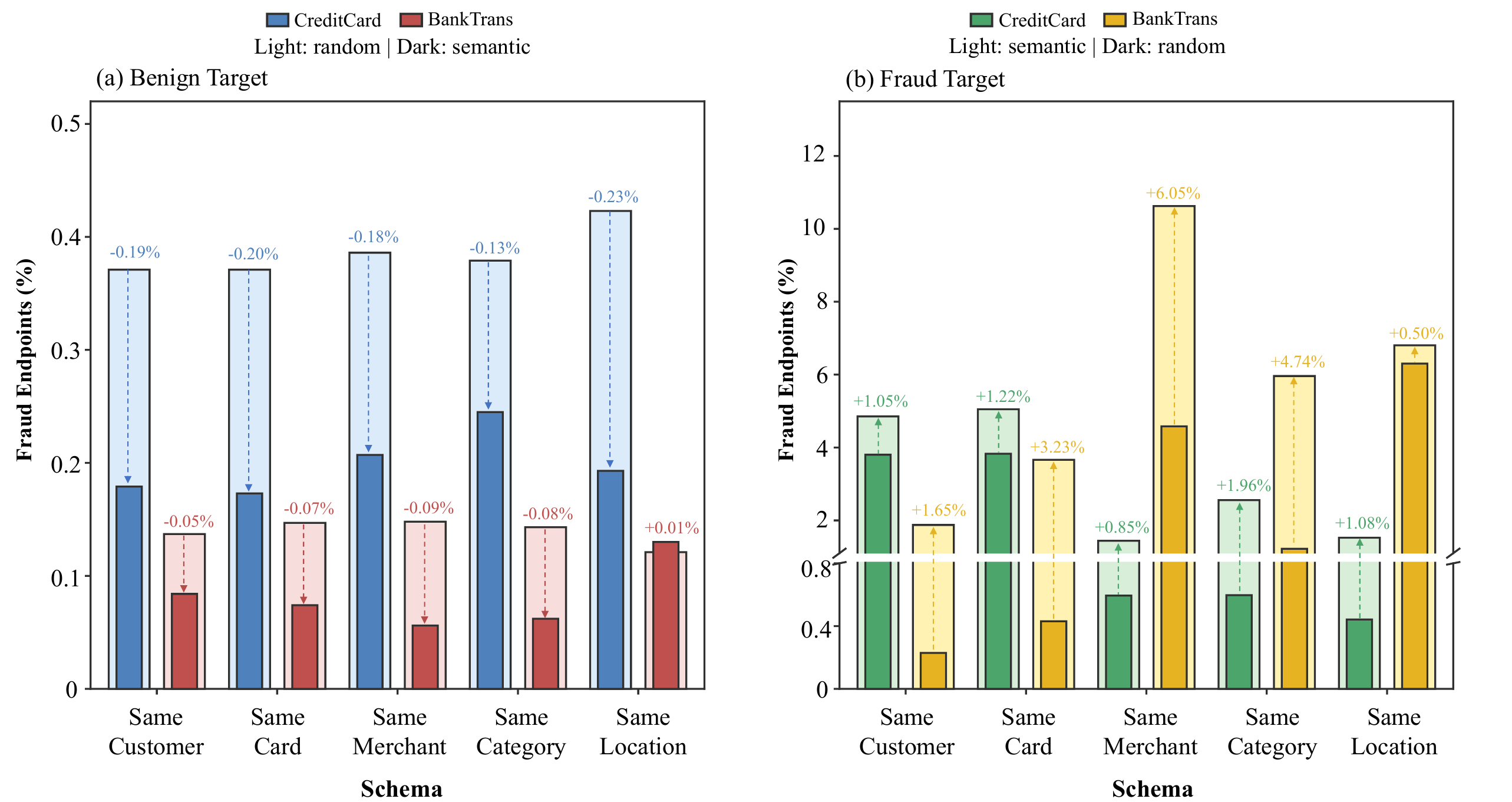}
    \caption{Fraud-label composition of sampled terminal transaction endpoints. Panels show benign and fraudulent targets; colors distinguish datasets, shades distinguish sampling strategies, and annotations report semantic-minus-uniform differences (percentage points).}
    \Description{Two panels compare fraud-endpoint rates under semantic-guided and uniform sampling across five transaction schemas for benign and fraudulent targets. Semantic-guided sampling generally lowers fraud-endpoint rates for benign targets and raises them for fraudulent targets.}
    \label{fig:endpoint-composition}
\end{figure}

Figure~\ref{fig:endpoint-composition} shows a consistent class-conditional pattern. For benign targets, semantic-guided sampling yields lower fraud-endpoint rates than uniform sampling for all five CreditCard schemas and four of the five BankTrans schemas. For the same-location schema on BankTrans, the two rates are very close. For fraudulent targets, semantic-guided sampling yields higher fraud-endpoint rates across all five schemas on both datasets. The size of this increase varies by schema: it is largest for same-category endpoints on CreditCard (+1.959 percentage points) and same-merchant endpoints on BankTrans (+6.046 points). Overall, the pattern is consistent with semantic guidance shifting the sampled evidence toward endpoints whose fraud-label composition aligns with the target class, while the endpoint labels themselves are not used by the sampler.

\section{Ethical Use of Data and Informed Consent}
\label{sec:ethics}

This study uses the publicly released CreditCard and BankTrans datasets from FinFraudBench~\cite{chen2026finfraudbench} and the Elliptic++ dataset~\cite{10.1145/3580305.3599803}. Our experiments are a secondary analysis of these existing transaction-graph data; we collected no new data and did not recruit or contact individuals.

\section{Conclusion and Future Work}
\label{sec:conclusion}

We presented SemHGT, an LLM-enhanced framework that makes role-aware business semantics explicit in heterogeneous transaction graphs. Its first stage generates risk-related business descriptions and encodes them as semantic vectors to enrich node representations; these vectors also guide target-specific sampling of typed path evidence. The sampled local evidence is then combined with global schema-neighbor context into schema-level tokens for transformer-based fraud prediction. Across CreditCard, BankTrans, and Elliptic++, SemHGT leads the evaluated baselines on most dataset--metric comparisons. Further analyses provide evidence for the effectiveness of the design and clarify how semantic modeling and evidence selection contribute to fraud detection. These findings support the value of combining role-aware semantics with relational evidence for financial fraud detection on heterogeneous transaction graphs.

Future work can investigate ways to reduce reliance on manually specified schemas, including methods that integrate role-aware business semantics into meta-path-free heterogeneous graph models for financial fraud detection. Another direction is to adapt semantic representations to fraud-detection objectives and evolving transaction behavior.

\clearpage
\bibliographystyle{ACM-Reference-Format}
\bibliography{references}

\clearpage
\appendix
\raggedbottom
\setlength{\textfloatsep}{2pt plus 1pt minus 1pt}
\setlength{\floatsep}{2pt plus 1pt minus 1pt}
\setlength{\intextsep}{5pt plus 1pt minus 1pt}
\section{Extended Experimental Details}
\label{app:experimental-details}
\subsection{Graph Views and Split Protocols}

Table~\ref{tab:split-counts} gives the transaction counts and class composition for the dataset splits described in Section~\ref{sec:experimental-setup}.

\begin{table}[!htbp]
\centering
\caption{Transaction-label counts for each data split.}
\label{tab:split-counts}
\footnotesize
\setlength{\tabcolsep}{4pt}
\begin{tabular}{llrrr}
\toprule
Dataset & Split & Transactions & Fraud & Non-fraud \\
\midrule
CreditCard & Train & 1,111,436 & 6,401 & 1,105,035 \\
 & Validation & 185,240 & 1,105 & 184,135 \\
 & Test & 555,718 & 2,145 & 553,573 \\
\midrule
BankTrans & Train & 891,496 & 1,333 & 890,163 \\
 & Validation & 891,496 & 1,333 & 890,163 \\
 & Test & 7,131,971 & 10,666 & 7,121,305 \\
\midrule
Elliptic++ & Train & 28,938 & 3,402 & 25,536 \\
 & Validation & 4,005 & 275 & 3,730 \\
 & Test & 13,621 & 868 & 12,753 \\
\bottomrule
\end{tabular}
\end{table}

\subsection{Validation and Full-Test Evaluation}

For validation, we retain all fraud transactions and subsample non-fraud transactions (Table~\ref{tab:validation-subsets}). We weight each sampled negative by the full-to-sampled negative-count ratio when computing AUPRC and AUROC. Checkpoint selection and F1-threshold calibration use validation labels; final metrics use the complete test split.

\begin{table}[!htbp]
\centering
\caption{Validation subsets used for checkpoint selection and threshold calibration.}
\label{tab:validation-subsets}
\scriptsize
\setlength{\tabcolsep}{3pt}
\begin{tabular}{@{}lrrr@{}}
\toprule
Dataset & Full validation & Subset (fraud / non-fraud) & Neg. weight \\
\midrule
CreditCard & 185,240 & 2,210 (1,105 / 1,105) & 166.6380 \\
BankTrans & 891,496 & 22,661 (1,333 / 21,328) & 41.7368 \\
Elliptic++ & 4,005 & 550 (275 / 275) & 13.5636 \\
\bottomrule
\end{tabular}
\end{table}

\subsection{Balanced Test Subsets for Analytical Experiments}
\label{app:analysis-subsets}

Except for ablation, analytical experiments use the fixed, class-balanced test subsets in Table~\ref{tab:analysis-subsets}.

\begin{table}[b]
\centering
\caption{Class-balanced test subsets used in analytical experiments other than ablation.}
\label{tab:analysis-subsets}
\scriptsize
\setlength{\tabcolsep}{4pt}
\begin{tabular}{@{}lrrr@{}}
\toprule
Dataset & Non-fraud & Fraud & Total \\
\midrule
CreditCard & 2,145 & 2,145 & 4,290 \\
BankTrans & 1,000 & 1,000 & 2,000 \\
Elliptic++ & 868 & 868 & 1,736 \\
\bottomrule
\end{tabular}
\end{table}

\subsection{Schema-Specific Sampling Budgets}

Table~\ref{tab:schema-budgets} lists per-schema path quotas and maximum context sizes. Sampling expands the transaction frontier for up to two rounds; the context cap counts typed intermediate entities.

\begin{table}[!htbp]
\centering
\caption{Local path schemas and sampling budgets. $T$ denotes a transaction; $W$ denotes a wallet.}
\label{tab:schema-budgets}
\scriptsize
\setlength{\tabcolsep}{3pt}
\begin{tabular}{@{}llrr@{}}
\toprule
Dataset & Schema & Per-round path quota $k_r$ & Max. context nodes \\
\midrule
\multirow{5}{*}{CreditCard} & $T$--Customer--$T$ & 8 & \multirow{5}{*}{112} \\
 & $T$--Card--$T$ & 8 & \\
 & $T$--Merchant--$T$ & 8 & \\
 & $T$--Category--$T$ & 1 & \\
 & $T$--Location--$T$ & 1 & \\
\midrule
\multirow{5}{*}{BankTrans} & $T$--Customer--$T$ & 8 & \multirow{5}{*}{128} \\
 & $T$--Card--$T$ & 8 & \\
 & $T$--Merchant--$T$ & 8 & \\
 & $T$--Category--$T$ & 4 & \\
 & $T$--Location--$T$ & 4 & \\
\midrule
\multirow{4}{*}{Elliptic++} & $T$--$T$ & 8 & \multirow{4}{*}{160} \\
 & $T$--$W$--$T$ & 8 & \\
 & $T$--$W$--$W$--$T$ & 4 & \\
 & $T$--$W$--$W$--$W$--$T$ & 4 & \\
\bottomrule
\end{tabular}
\end{table}

\subsection{Fraud-Label Composition of Sampled Evidence}
\label{app:endpoint-composition}

Table~\ref{tab:endpoint-composition} reports fraud rates among unique terminal transaction endpoints, excluding the target, for semantic-guided and uniform sampling on the balanced subsets in Table~\ref{tab:analysis-subsets}. Endpoint labels are used only for this post-sampling analysis.

\begin{table}[H]
    \centering
    \caption{Fraud-endpoint rates for semantic-guided and uniform sampling. Rates are percentages; $\Delta$ is semantic minus uniform in percentage points.}
    \label{tab:endpoint-composition}
    \scriptsize
    \setlength{\tabcolsep}{2.5pt}
    \begin{tabular}{@{}llrrl@{}}
        \toprule
        Target label & Schema & Semantic & Uniform & $\Delta_{\mathrm{sem-unif}}$ \\
        \midrule
        \multicolumn{5}{@{}l}{\textbf{CreditCard}} \\
        \addlinespace[2pt]
        Benign & same customer & 0.179 & 0.371 & -0.192 \\
        Benign & same card & 0.173 & 0.371 & -0.198 \\
        Benign & same merchant & 0.207 & 0.386 & -0.179 \\
        Benign & same category & 0.245 & 0.379 & -0.135 \\
        Benign & same location & 0.193 & 0.423 & -0.230 \\
        Fraudulent & same customer & 4.854 & 3.803 & +1.051 \\
        Fraudulent & same card & 5.050 & 3.829 & +1.221 \\
        Fraudulent & same merchant & 1.439 & 0.594 & +0.845 \\
        Fraudulent & same category & 2.556 & 0.597 & +1.959 \\
        Fraudulent & same location & 1.524 & 0.442 & +1.082 \\
        \midrule
        \multicolumn{5}{@{}l}{\textbf{BankTrans}} \\
        \addlinespace[2pt]
        Benign & same customer & 0.084 & 0.137 & -0.054 \\
        Benign & same card & 0.074 & 0.147 & -0.072 \\
        Benign & same merchant & 0.056 & 0.148 & -0.092 \\
        Benign & same category & 0.062 & 0.143 & -0.080 \\
        Benign & same location & 0.130 & 0.121 & +0.009 \\
        Fraudulent & same customer & 1.875 & 0.229 & +1.646 \\
        Fraudulent & same card & 3.663 & 0.431 & +3.232 \\
        Fraudulent & same merchant & 10.626 & 4.580 & +6.046 \\
        Fraudulent & same category & 5.962 & 1.223 & +4.738 \\
        Fraudulent & same location & 6.805 & 6.305 & +0.500 \\
        \bottomrule
    \end{tabular}
\end{table}

\FloatBarrier
\section{CreditCard Risk-Semantic Prompt Templates}
\label{app:creditcard-prompt}

The CreditCard prompts pair a shared system message with a node-type-specific instruction for each of the six node types.

\subsection{Shared System Message}
\label{app:creditcard-prompt-template}

\begin{promptbox}{System message}
You are a financial graph semantic analyst supporting transaction fraud-risk modeling. Return exactly one line in this format and no other text: Business risk level: <low|moderate|high>. Evidence: <one complete factual sentence of at most 32 English words>. Choose the level only from supplied attributes and supplied reference ranges. When available, the evidence should cover two or three complementary observable fields, such as amount or aggregate, decoded category, time, geography, or behavior. Do not compare with percentiles or ranges unless the comparison is explicitly and unambiguously supported. Do not state a fraud or benign label, use labels, data splits, entity identifiers, causal claims, or invented facts.
\end{promptbox}

\subsection{Node-type-specific User Instructions}
\label{app:creditcard-node-prompts}

\begin{promptbox}{Transaction}
Describe the payment event only from its supplied amount, local timestamp or weekend context, customer-versus-merchant distance, coordinates, and local population. Treat a large amount or long distance as context only, never as a label.
\end{promptbox}

\begin{promptbox}{Customer}
Describe the customer activity profile using age, home geography, transaction count, total and typical spend, spend variability, activity duration, and card, merchant, and category diversity. Use relative ranges when supplied.
\end{promptbox}

\begin{promptbox}{Card}
Describe the card usage profile using transaction volume, total, typical, variable, maximum and minimum spend, active duration, billing state, and merchant/category diversity. State relative exposure only when supported by the reference range.
\end{promptbox}

\begin{promptbox}{Merchant}
Describe the merchant context using business category, transaction scale, typical and variable amount, customer/card reach, and state diversity. Do not use the merchant name as evidence.
\end{promptbox}

\begin{promptbox}{Category}
Describe the business-category context using transaction scale, amount profile, merchant coverage, and card/customer reach relative to category-level reference ranges.
\end{promptbox}

\begin{promptbox}{Location}
Describe the geographic activity context using transaction scale, amount profile, customer/card/merchant reach, and population context relative to location-level ranges.
\end{promptbox}

The selected instruction is paired with the shared system message, and filtered node attributes are supplied as name--value lines. When available, prompts also include label-blind reference bands and summary statistics. Permitted fields and reference statistics depend on node type; CreditCard uses training-only statistics for training nodes, training-plus-validation statistics for validation nodes, and training-plus-test statistics for test nodes.

\end{document}